\documentclass[10pt,twocolumn,letterpaper]{article}

\usepackage{iccv}
\usepackage{times}
\usepackage{epsfig}
\usepackage{graphicx}
\usepackage{subcaption}
\usepackage{amsmath}
\usepackage{amssymb}
\usepackage{diagbox}
\usepackage{rotating}
\usepackage{algorithm}
\usepackage{algorithmic}
\usepackage{dblfloatfix}
\usepackage{placeins}

\usepackage{hyperref}
\hypersetup{pagebackref=true,breaklinks=true,letterpaper=true,colorlinks,bookmarks=false}

\iccvfinalcopy

\def\iccvPaperID{****}

\ificcvfinal\fi

\begin{document}

%%%%%%%%% TITLE
\title{RMOPP: Robust Multi-Objective Post-Processing for Effective Object Detection}

\author{Mayuresh Savargaonkar \qquad Abdallah Chehade\\
Department of Industrial and Manufacturing Systems Engineering\\
University of Michigan-Dearborn\\
\and Samir Rawashdeh\\
Department of Electrical and Computer Engineering\\
University of Michigan-Dearborn\\
\tt\small mayuresh@umich.edu \qquad \tt\small achehade@umich.edu \qquad \tt\small srawa@umich.edu}
\maketitle
.
\ificcvfinal\thispagestyle{plain}\fi

%%%%%%%%% ABSTRACT
\begin{abstract}
Over the last few decades, many architectures have been developed that harness the power of neural networks to detect objects in near real-time. Training such systems requires substantial time across multiple GPUs and massive labeled training datasets. Although the goal of these systems is generalizability, they are often impractical in real-life applications due to flexibility, robustness, or speed issues. This paper proposes RMOPP: A robust multi-objective post-processing algorithm to boost the performance of fast pre-trained object detectors with a negligible impact on their speed. Specifically, RMOPP is a statistically driven, post-processing algorithm that allows for simultaneous optimization of precision and recall. A unique feature of RMOPP is the Pareto frontier that identifies dominant possible post-processed detectors to optimize for both precision and recall. RMOPP explores the full potential of a pre-trained object detector and is deployable for near real-time predictions. We also provide a compelling test case on YOLOv2 using the MS-COCO dataset.
\end{abstract}

%%%%%%%%% BODY TEXT
\section{Introduction}

Object detectors and vision systems have greatly advanced in the last few years. The evolution of deep neural networks has unlocked new potential in complex fields such as capacity estimations of Li-ion battery cells \cite{Shi2020,Savargaonkar2020,Chehade2019c} and Remaining-Useful-Life (RUL) predictions \cite{Shi2021,Chehade2017,Chehade2019}. The field of object detection and classification is another significant beneficiary of this especially due to the advancements in Convolutional Neural Networks (CNNs). In this paper, we refer to the broader task of detection and classification simply as object detection. Deep convolution neural networks commonly outperform traditional methods such as ones using Support Vector Machines (SVMs) or regression techniques \cite{Liu2019}. CNNs are typically either run as a single-stage detector or a two-stage detector. More details on state-of-art detectors are provided in the literature review.

Research today mainly focuses on building object detection systems that offer a reasonable trade-off between speed and accuracy. For improved accuracy, two-stage detectors including ensemble neural networks like PA Net \cite{Liu2018} are shown to perform well in competitions like PASCAL VOC \cite{M.2010}, MS-COCO \cite{Lin2014}, and ImageNet \cite{Deng2010}. Although such systems provide competition winning performances, they sacrifice speed and are not currently suitable for deployment in real-time systems. For improved latency, single-stage detectors like the  YOLO series algorithms: YOLOv1 \cite{Redmon2016}, YOLOv2 \cite{Redmon2017}, and YOLOv3 \cite{Farhadi2018} are shown to be faster than two-stage detectors. Although single-stage detectors are extremely fast, they often fail to reach the same level of robustness as the two-stage detectors \cite{Liu2019}. Robustness here is defined as the ability of an object detection pipeline to successfully handle perturbations. Single-stage detectors’ limitations are often associated with localization errors and the identification of false positives \cite{Puttemans2018}. This makes them less suitable for critical applications that require high robustness.

\begin{figure}[t]
\centering
\begin{subfigure}[t]{\linewidth}
\centering
\includegraphics[width=0.49\linewidth,height=1.2in]{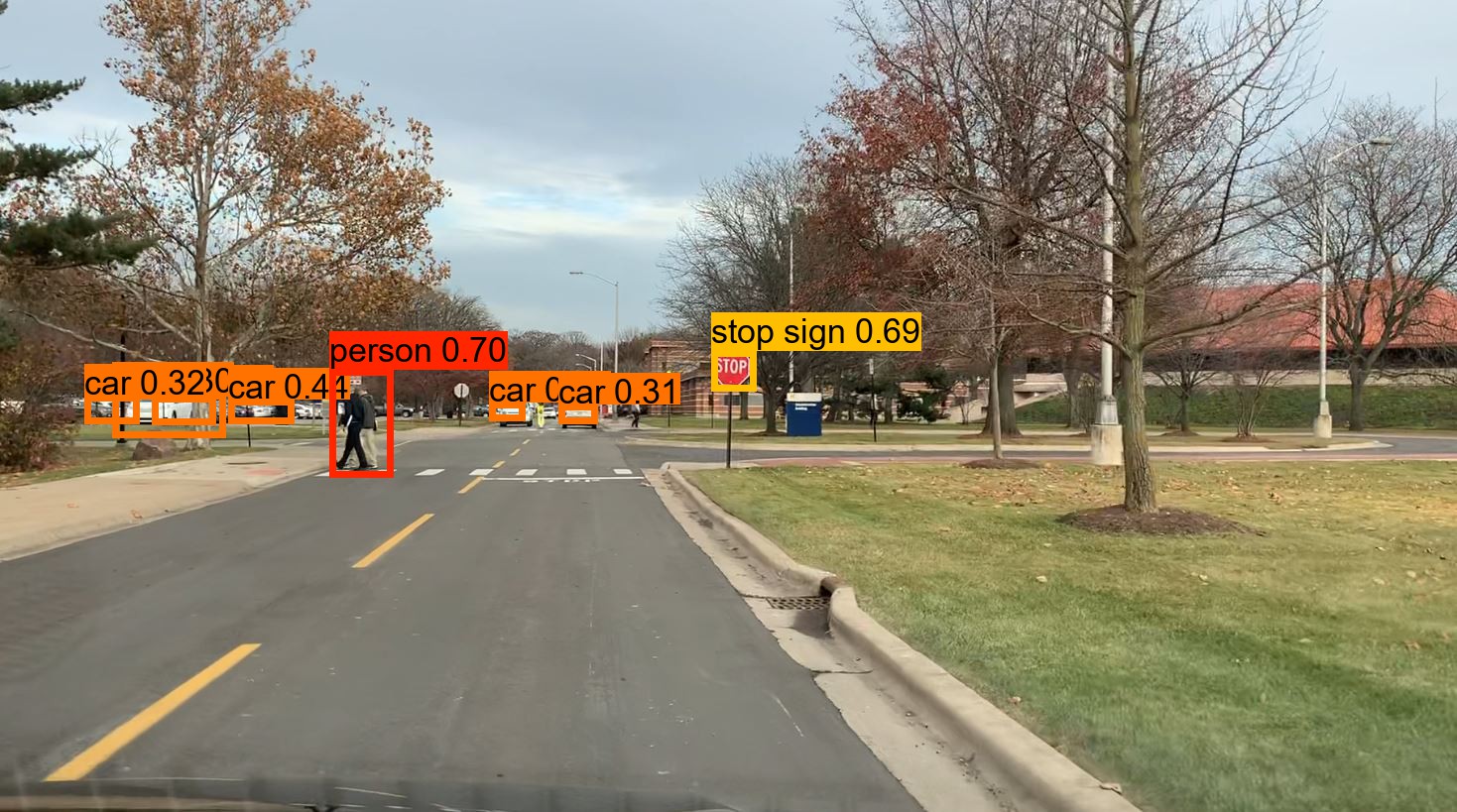} 
\includegraphics[width=0.49\linewidth,height=1.2in]{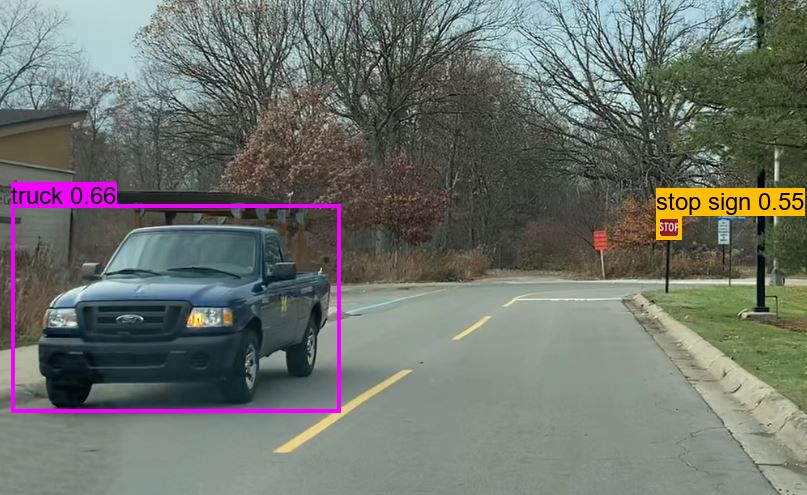}
\end{subfigure}
\caption{\label{fig:fig1}Example of object detection in modern vehicles.}
\end{figure}

Practically, for many real-life applications like object detection in autonomous vehicles in Figure \ref{fig:fig1}, both, detection of an object and speed are deemed extremely important. A failure in detection, as well as a slow detection, can result in a fatal crash. Further, these object detection pipelines are often hard-wired to achieve high precision at lower recall or higher recall at lower precision to achieve competition winning performances. Therefore, there is an essential need for a customizable and low-latency object detection pipeline with a robust post-processing method that can process at real-time speeds without undergoing re-training. 

Generally, it is hard to tune a robust object detection pipeline with a uni-objective post-processing algorithm. While existing methods focus on optimizing either precision/recall they suffer from maintaining an acceptable recall/precision.  In this paper, we propose RMOPP: A robust multi-objective post-processing algorithm that allows for simultaneous optimization of both, precision and recall. RMOPP exhibits the following advantages:

\begin{itemize}
	\item {\bf Statistical Guarantees}- Development of a heuristic based on the log-likelihood ratio provides explainable statistical reasoning.
	\item {\bf Increased Robustness}- Intuitive post-processing hyper-parameters developed using statistical guarantees allow for robust and improved filtering.
	\item {\bf Pareto Frontier}- Development of a multi-dimensional post-processing algorithm allows for a thorough exploration of an object detection pipeline using the concepts of Pareto frontier. This helps in the choice of optimized post-processing hyper-parameters. More details on the Pareto frontier are discussed in Section \ref{section:4}.
	\item {\bf Need for Re-training}- The need for re-training the object detector is eliminated when preference-based optimization for Precision and/or Recall is required.
	\item {\bf Decreased Complexity}- Avoids design, development, and training of complex post-processing procedures while achieving state-of-art performances.
	\item {\bf Low Latency}- Since the post-processing hyper-parameters are optimized offline, the inference time of the object detection pipeline remains unaffected.

\end{itemize}

Although we present a case study based on the single-stage detectors in this paper, there is no reason to assume that a similar change in the post-processing step won’t be beneficial for the two-stage detectors. The remainder of the paper is organized as follows: Section \ref{section:2} explores literature in the field of object detection and post-processing, Section \ref{section:3} introduces the proposed post-processing algorithm, Section \ref{section:4} verifies the efficacy of the proposed algorithm using a case-study and Section \ref{section:5} concludes the paper with directions for future work.

\section{Literature Review}\label{section:2}

\begin{figure*}
\centering
\includegraphics[width=0.9\linewidth]{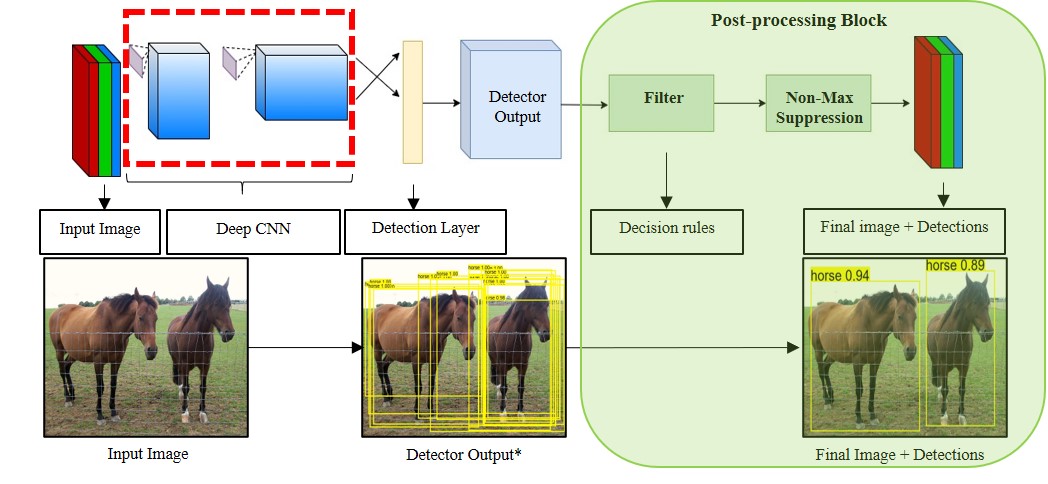}
\caption{Object detection pipeline using a single-stage detector. *Detector output is cleaned for visualization purposes.}
\label{fig:fig2}
\end{figure*}

\subsection{Object Detectors: State-of-Art}
Modern object detectors can be categorized into two types: (i) Two-stage detectors like R-CNN \cite{Girshick2014}, Fast R-CNN \cite{Girshick2015}, Faster R-CNN \cite{Ren2016}, and many others, (ii) Single-stage detectors like OverFeat \cite{Sermanet2014}, YOLO \cite{Redmon2016}, RetinaNet \cite{Lin2020} and many others. The family of R-CNN type two-stage object detectors works on selective search strategies or Region Proposal Networks (RPN) \cite{Girshick2014}. The first stage proposes and filters locations inside the image that have a high probability of containing an object, and the second stage scores these proposed regions. Fast R-CNN reduces the computational load in the first stage by passing the image only once through a convolutional layer that generates Regions of Interest (ROI) using the selective search strategy. Faster R-CNN improves upon Fast R-CNN by using an RPN instead of a selective search strategy which further offers a 10x improvement in speed \cite{Ren2016}. Cascade R-CNN \cite{Cai2018} uses multiple R-CNN modules capable of detecting varying sizes of objects with different classification thresholds for each module. Frameworks such as Libra-RCNN \cite{Pang2019} work on performance improvements by working on the imbalance in training images. To offer speed improvements over RPN stages of Fast R-CNN, a novel method, CPNDet \cite{Duan2020} was recently published by Duan \etal. CPNDet replaces the RPN stage with an anchor-free heat-map based detection framework called as CornerNet \cite{Law2018}. Although two-stage detectors show better detection performance, they offer limited improvements in speed due to several complex modules being stitched together to produce final detections. Moreover, these detectors have far too many hyper-parameters which make them unsuitable for real-life applications. 

Single-stage detectors are fundamentally different and try to achieve the entire process of detection, classification, and bounding box regression using a single step. Most single-stage detectors work extremely fast but fail to reach the same level of robustness as two-stage detectors.  Algorithms such as DSSD \cite{Fu2017}, SSD \cite{Liu2016}, and YOLO \cite{Redmon2017}, \cite{Farhadi2018} use anchor-based multi-scale detections for identifying fine-grained features in the image and thus improve the accuracy of the detector. Object detectors like RetinaNet \cite{Lin2020} try to improve on the localization errors created by the single-stage detectors by introducing  ‘focal loss’ that rewards the model under training when it correctly classifies the hard-negative examples. To eliminate problems associated with detections of small-scale objects in common single-stage detectors, researchers recently introduced detectors such as  CenterNet \cite{Duan2019} and FCOS \cite{Tian2019}. These detectors work by eliminating the need for anchor boxes. In their detailed experiments with ATSS, Zhang \etal \cite{Zhang2020} show that definitions of positive and negative annotations help bridge traditional gaps between anchor-free and anchor-based methods.  

End-to-end object detection is another branch of single-stage detectors that try to improve performance by eliminating the post-processing stages such as NMS (Non-Maximal Suppression). DETR \cite{Carion2020} is the first such detector that is capable of performing end-to-end object detection. OneNet \cite{Sun2020} is another recent end-to-end object detector that achieves state-of-art while running at real-time speeds. While end-to-end object detectors achieve unparalleled performances in speed and accuracy, they lack customizability offered by post-processing stages. Customizations require re-training using several GPUs and hours of complex training paradigms often unavailable to the common user. The importance of this subject is also touched upon by J. Huang \etal \cite{Huang2017a}, although they focus more on exploiting the aspect of speed and accuracy trade-offs for developing a better detector. While investigating the effect of speed on accuracy for a wide range of detectors, they show that the single-stage detectors based on the ResNet 101 backbone perform equally well when compared to R-FCN \cite{Huang2017a} and Faster RCNN on the same backbone for large object detection on the MS-COCO dataset.

\subsection{Post-Processing Algorithms: State-of-Art}

Most existing object detectors start with outputting four bounding box co-ordinates, a score, and the class associated with each one of the detections. The commonly used ‘score’ function for detection is the product of the probability of the most probable class bounded in an identified detection and the confidence of having an object inside the detected box. Next, detections with low scores are filtered assuming that lower scores represent poorly identified or classified objects. Once the detections are filtered, NMS is applied with a pre-defined NMS threshold ($\eta$) \cite{Bodla2017}. NMS is the process of identifying detections with the Intersection over Union ($\text{IoU}$) greater than the NMS threshold ($\eta$) and then eliminating those with lower ‘scores’. The NMS process is effective in eliminating redundant detections.

A significant improvement in post-processing comes as the ‘Soft-NMS’ replacing the traditional greedy NMS \cite{Bodla2017}. This work has shown to be of little use in the case of complex datasets like the PASCAL VOC and MS-COCO \cite{Bodla2017}. To propose a suitable improvement over the ‘Soft-NMS’, He \etal \cite{He2019} introduced the ‘Softer-NMS’ which decays the bounding box scores using a continuous function thus achieving better results. Relation networks \cite{Hu2018} and Learning NMS \cite{Hosang2017} are amongst some other types of works that try to improve upon the traditional NMS by designing and training a sub-network to analyze complex object-object correlations. While works such as Fitness NMS \cite{Tychsen-Smith2018} try to integrate localization information into ranking scores, LTR \cite{Tan2019} is another sub-network that improves on suppression ranking via learning procedures. NMS based post-processing works show promising performance gains and thus help build confidence that effective post-processing is key in any suitable object detection pipeline. A summary of a common single-stage object detection pipeline is shown in Figure \ref{fig:fig2}.

Although there exist post-processing algorithms, they often fail to explicitly acknowledge the fact that there may exist multiple recalls at the same precision and vice-versa. Thus, they fail to explore the full potential of an object detection pipeline which results in sub-optimal detection performance. RMOPP thus uses the concepts of Pareto frontier which identifies a set of dominant post-processing hyper-parameters that optimize both precision and recall simultaneously. More details are provided in Section \ref{section:4}.

\section{Proposed Algorithm: RMOPP}\label{section:3}

Existing post-processing algorithms rely on a single data-driven ‘score’ metric that is not statistically intuitive. Several assumptions are made when filtering detections by scores and NMS using pre-defined post-processing hyper-parameters and thresholds. These values are usually chosen on an empirical basis, user’s experience, or by using a trial-and-error method. Hereafter, we will focus on the most common ‘score’ function for existing post-processing algorithms that is defined by the product of the probability of the most probable class bounded in an identified detection and the confidence of having an object inside the detected bounding box. 

With this definition of the score function, Table \ref{table:table1} summarizes the output of commonly used object detectors. Block I comprise detections with poor scores that are typically filtered out. Block IV comprises of detections with high scores that are typically retained. Block II and III comprise of risky detections with intermediate scores. The existing post-processing algorithms are susceptible to poor performance for detections in Block II and III that have intermediate level scores. Many of those detections are (i) mistakenly filtered out (a large number of false negatives) thus achieving a lower than ideal recall or, (ii) poor detections are retained (a large number of false positives) thus achieving a lower than ideal precision. To address this limitation, we introduce RMOPP: A robust multi-objective, post-processing algorithm that allows simultaneous and optimized tuning of recall and precision.

\begin{table}
\begin{center}
\begin{tabular}{|l|c|c|}\hline
\diagbox[width=5cm]{Probability of\\ most probable class}{Probability of\\ existence of\\ bounding box}&{Low}&{High}\\\hline\hline
{Low}&{\bf Block I}&{Block II}\\
{High} &{Block III}&{\bf Block IV}\\\hline
\end{tabular}
\end{center}
\caption{\label{table:table1}The blocks in {\bf ‘bold’} show where the existing post-processing algorithms are effective.}
\end{table}

\subsection{Proposed Post-Processing Algorithm}
Let $(X^c)$  be the binary random variable associated with the existence of some bounding box $(c)$ with probability $(P_c)$. Let $({{X}_{i}}^{c})$ be the binary random variable indicating that the object in the bounding box $(c)$ belongs to a class $(i)$ with probability  $({{P}_{i}}^{c})$. Under the assumptions that ${{X}^{c}}\sim ~Ber({{P}_{c}})$ and the collection $\left\{ {{X}_{1}}^{c},...,{{X}_{i}}^{c},...,{{X}_{N}}^{c} \right\}\sim MultiNom\left( {{P}_{i}}^{c},...,{{P}_{N}}^{c},1 \right)$,
\begin{equation}
E[{{X}^{c}}]=P({{X}^{c}}=1)={{P}_{c}}
\end{equation}

\begin{equation}
E[X_{i}^{c}]=P(X_{i}^{c}=1)=P_{i}^{c}
\end{equation}

\begin{equation}
\sum\limits_{i=1}^{N}{P_{i}^{c}=1}
\end{equation}

To simplify the mathematical notations, let $({{Z}_{i}}^{c})$ be the ordered statistic of $({{X}_{i}}^{c})$ such that $P(Z_{i}^{c}=1)\text{ x }P(Z_{j}^{c}=1)\text{ }\forall \text{ }j>i$. Under this notation, $P(Z_{i}^{c}=1)$ is the probability of the most probable class for the detection bounded in the box $(c)$ and the existing post-processing technique is expressed as,

\begin{equation} \label{eq:4}
{{s}^{c}}=P(Z_{1}^{c}=1)\text{ x }P({{X}^{c}}=1)\le \gamma 
\end{equation}

\noindent where the detection in the box $(c)$ is filtered out if its score $(s^c)$ is less than a pre-defined threshold $(\gamma)$ as stated in Equation \ref{eq:4}. The threshold $(\gamma)$ is often chosen based on empirical analysis, user experience, or by limited trials. Extensions of the of Equation \ref{eq:4} exist where the score is defined as $\underset{i}{\mathop{\max }}\,P({{X}_{i}}^{c}=1)$ or as $\underset{i}{\mathop{\max }}\,P({{X}}^{c}=1)$. It should be noted that if the object detector does not output the confidence $P({{X}}^{c}=1)$, then $P({{X}}^{c}=1)$ is set to 1 for all detections. 

Note that during training, object detectors aim to minimize the population cross-entropy over the training dataset. Therefore, it is natural to design a post-processing algorithm that shares similar merit to the training procedure. Specifically, we consider thresholding over the log-likelihood ratios of the most probable class to all other classes for every bounding box. This brings us to our first proposition.

\medskip
\noindent{\bf Proposition I}: The log-likelihood ratios between the top class and the remaining classes of a bounding box $(c)$ are sufficient to quantify the classification accuracy of the object in the bounding box $(c)$.

Proposition I is inspired by the likelihood ratio test and the theory of statistical hypothesis testing, which shows that every bounding box $(c)$ must satisfy $\left( \frac{P(Z_{1}^{c}=1)}{P(Z_{k}^{c}=1)} \right)\ge {{\gamma }_{1}},\text{ }\forall k>1$.

\medskip
\noindent{\bf Lemma I}: Given the definition that $({{Z}_{i}}^{c})$ is the ordered statistic of $({{X}_{i}}^{c})$ such that $P(Z_{i}^{c}=1)\text{ x }P(Z_{j}^{c}=1)\text{ }\forall\text{ }j>i$. Then,$\left( \frac{P(Z_{1}^{c}=1)}{P(Z_{2}^{c}=1)} \right)\ge {{\gamma }_{1}}$ for any bounding box $(c)$ guarantees  $\left( \frac{P(Z_{1}^{c}=1)}{P(Z_{k}^{c}=1)} \right)\ge {{\gamma }_{1}},\text{ }\forall k>1$.
	From Lemma I and Proposition I, we identify our first score metric to be $\left( \frac{P(Z_{1}^{c}=1)}{P(Z_{2}^{c}=1)} \right)$ and its user-defined threshold to be $(\gamma_1)$. Therefore, any detected box $(c)$ must satisfy Equation \ref{eq:5} or it will be removed for poorly classifying its bounded object.

\begin{equation} \label{eq:5}
\left( \frac{P(Z_{1}^{c}=1)}{P(Z_{2}^{c}=1)} \right)\ge {{\gamma }_{1}},\text{ }\forall c
\end{equation}

While the classification task is critical in object detection, the bounding box detection is also an important task. Proposition I guarantees a good classification but it does not guarantee a good bounding box detection. This brings us to Proposition II.

\medskip
\noindent{\bf Proposition II}: An effective object detector must show a relatively similar classification and detection performance which can be quantified by the log-likelihood ratio of detection to classification for any bounding box $(c)$, $\left( \frac{P({{X}^{c}}=1)}{P(Z_{1}^{c}=1)} \right),\text{ }\forall k>1$.

Proposition II suggests that a detected bounding box truly bounds an object if $\left( \frac{P({{X}^{c}}=1)}{P(Z_{1}^{c}=1)} \right),\text{ }\forall k>1$, is large enough. It should be noted here that there exists a cross-correlation between Propositions I and II, where Proposition II mainly focuses on the relative detection performance given the classification probabilities $P\{({{X}^{c}}=1)|P(Z_{1}^{c}=1),...,P(Z_{N}^{c}=1)\}$ for any detected bounding box $(c)$, thus ensuring that the boxes are well-identified and classified.

\medskip
\noindent{\bf Lemma II}:  Given the definition that $({{Z}_{i}}^{c})$ is the ordered statistic of $({{X}_{i}}^{c})$ such that $P(Z_{i}^{c}=1)\text{ x }P(Z_{j}^{c}=1)\text{ }\forall\text{ }j>i$. Then, $\left( \frac{P({{X}^{c}}=1)}{P(Z_{1}^{c}=1)} \right)\ge {{\gamma }_{2}}$ for any bounding box $(c)$, guarantees  $\left( \frac{P({{X}^{c}}=1)}{P(Z_{k}^{c}=1)} \right)\ge {{\gamma }_{2}},\text{ }\forall k>1$. From Lemma II and Proposition II, we identify our second score metric to be $\left( \frac{P({{X}^{c}}=1)}{P(Z_{1}^{c}=1)} \right)$ and its user-defined threshold to be $(\gamma_2)$. Therefore, any detected box $(c)$ must satisfy Equation \ref{eq:6} or it will be considered a false detection and removed.

\begin{equation} \label{eq:6}
\left( \frac{P({{X}^{c}}=1)}{P(Z_{1}^{c}=1)} \right)\ge {{\gamma }_{2}},\text{ }\forall c
\end{equation}

Both Equations \ref{eq:5} and \ref{eq:6} summarize the proposed post-processing algorithm, where a bounding box $(c)$ is a good detection if and only if,

\begin{equation} \label{eq:7}
\left( \frac{P(Z_{1}^{c}=1)}{P(Z_{2}^{c}=1)} \right)\ge {{\gamma }_{1}}\text{ }\cap \text{ }\left( \frac{P({{X}^{c}}=1)}{P(Z_{1}^{c}=1)} \right)\ge {{\gamma }_{2}},\text{ }\forall c
\end{equation}

\begin{figure}%
\centering
\begin{subfigure}[t]{0.5\textwidth}
\centering
{\includegraphics[height=1.5in,width=1.2in]{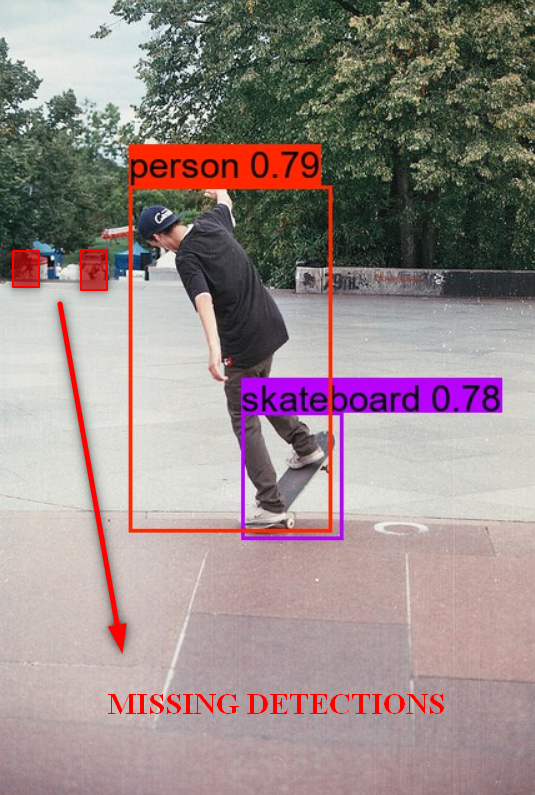} }
\hspace{0.75em}%
{\includegraphics[height=1.5in,width=1.2in]{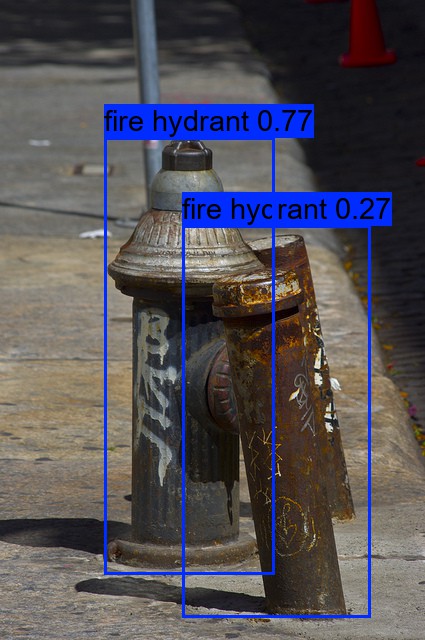} }
\caption{{\bf Before} using RMOPP}
\end{subfigure}
\begin{subfigure}[t]{0.5\textwidth}
\centering
{\includegraphics[height=1.5in,width=1.2in]{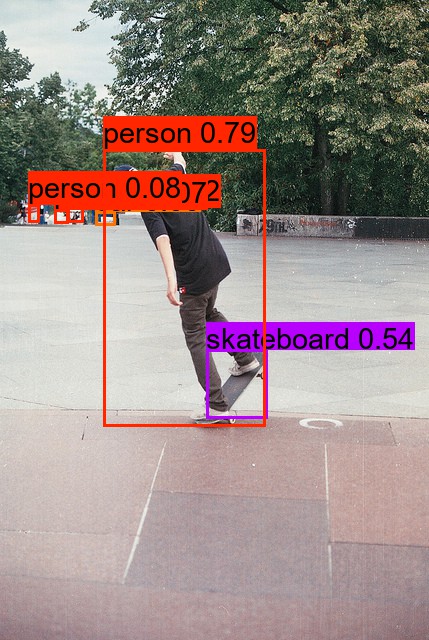} }
\hspace{0.75em}%
{\includegraphics[height=1.5in,width=1.2in]{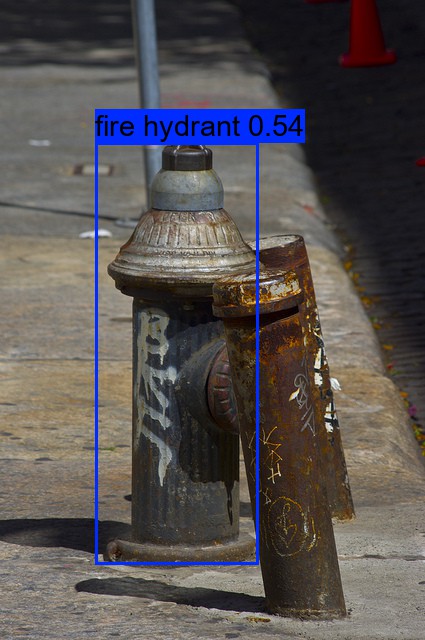} }
\caption{{\bf After} using RMOPP}
\end{subfigure}
\caption{$(\gamma_1 )$ and $(\gamma_2 )$ enable RMOPP to perform effective post-processing while eliminating false positives.}%
\label{fig:fig3}%
\end{figure}

\subsection{Advantages of RMOPP}

Equation \ref{eq:5} reduces the chances of having false positives due to poor classification. This addresses the problems with Block III in Table \ref{table:table1} for traditional post-processing algorithms because it filters out detections with low  $\left( \frac{P({{X}^{c}}=1)}{P(Z_{1}^{c}=1)} \right)$. Note that a lower value for  $\left( \frac{P({{X}^{c}}=1)}{P(Z_{1}^{c}=1)} \right)$ indicates poor classification based on Lemma I and Proposition I. Thus, it is hypothesized that increasing the value $(\gamma_1)$ will reduce false positives due to poor classification. Note that extreme values for $(\gamma_1)$ may have a detrimental effect on true positives, thus increasing the false negatives and lowering precision and hence, should be chosen carefully.  More details and results are presented in Section \ref{section:4}. 

Further, Lemma II helps address problems due to cases following Block II in Table \ref{table:table1}. Intuitively, $P(Z_1^c=1)$, should only be considered when $P(X^c=1)$ is high enough. At random detections where $P(X^c=1)$ is low and $P(Z_1^c=1)$ is very high, we can observe that the traditional algorithms fail as they eliminate too many true positives. Thus, it is hypothesized that a lower $(\gamma_2)$ will help increasing true positives by keeping detections that were traditionally eliminated due to poor identification. The cross-correlation between $(\gamma_1)$ and $(\gamma_2)$ will be particularly helpful here in keeping the false positives at bay. Figure \ref{fig:fig3} shows how results improve when RMOPP is used for filtering the detections compared to the traditional algorithms.

\subsection{Selecting Hyper-Parameters $(\gamma_1)$ and $(\gamma_2)$}

Choosing  $(\gamma_1)$ and $(\gamma_2)$ is a non-trivial task that depends on the object detector’s raw output. This is further elaborated in the case-study. Generally, decreasing both $(\gamma_1)$ and $(\gamma_2)$ will result in filtering out fewer detections. In extreme cases, this may result in poor precision due to extreme under-filtering. Similarly, increasing both $(\gamma_1)$ and $(\gamma_2)$ will result in filtering out increasing detections. In extreme cases, this may result in poor recall due to extreme over-filtering. We recommend that the user first decide on the evaluation metric of interest. The common three metrics we discuss in this paper are Precision, Recall, and $\text{F}_1$ score. Once decided, we apply Algorithm 1 with varying values of $(\gamma_1)$ and $(\gamma_2)$ to fully understand the complex correlations between the evaluation metric and $(\gamma_1)$ and $(\gamma_2)$. The Pareto frontier is then identified from these values which helps optimize for recall and precision simultaneously. It is worth noting that although there may exist multiple combinations of $(\gamma_1)$ and $(\gamma_2)$ that result in the same precision (or recall) but different recalls (or precisions) only the dominant combinations will be identified by the Pareto frontier. More details are provided in Section \ref{section:4} (Figure \ref{fig:fig4} and Figure \ref{fig:fig5}). Unlike existing post-processing algorithms, RMOPP is multi-objective and hence provides the utmost flexibility in simultaneously optimizing for precision and recall.

\begin{algorithm}[!t]
\caption{Pseudocode for RMOPP}
\begin{algorithmic}
\REQUIRE {$\left\{ \underset{i}{\mathop{\text{argmax}}}\,\text{ }{{X}_{i}}^{c},Z_{1}^{c},Z_{2}^{c},{{X}^{c}},{{b}^{c}} \right\}_{c=1}^{M}$}
\STATE $M$ predictions for a given image with bbox coord $b^c$;
\STATE $\delta_1$,$\delta_2$ are the increments for $\gamma_1$, $\gamma_2$;

\FOR {$\Delta _{2}^{L}\le\gamma_2\le\Delta _{2}^{U}$}
	\FOR {$\Delta _{1}^{L}\le\gamma_1\le\Delta _{1}^{U}$}
		\STATE {$I={0}^M$; an indicator if the detections satisfy Eq. \ref{eq:7}}
		\FOR {$c=1,...,M$}
			\IF {$\left( \frac{P(Z_{1}^{c}=1)}{P(Z_{2}^{c}=1)} \right)\ge {{\gamma }_{1}}\text{ }\text{and} \text{ }\left( \frac{P({{X}^{c}}=1)}{P(Z_{1}^{c}=1)} \right)\ge {{\gamma }_{2}}$}
				\STATE {$I[c]=1$}
			\ENDIF
		\ENDFOR
		\IF {$\sum (I)\ne 0$}
			\STATE {{\bf apply} NMS with $\text{IoU}=\eta$}
			\STATE {{\bf compute} $Precision, Recall \text{ and } F_1$}
		\ENDIF
		\STATE {$\gamma_1=\gamma_1 + \delta_1$}
	\ENDFOR 		
\STATE {$\gamma_2=\gamma_2+\delta_2$}
\ENDFOR
\RETURN {Precision, Recall and $\text{F}_1$ scores $\forall$ $\gamma_1$ and $\gamma_2$ }
\end{algorithmic}
\end{algorithm}

In summary, the object detection pipeline with RMOPP performs the following steps: (1.) Take an input image and pass it through the pre-trained detector that outputs some detections, (2.) Filter these detections using Equation \ref{eq:7}, given some values for $(\gamma_1)$ and $(\gamma_2)$ (3.) Perform NMS on detections from step 2 to further refine the results using NMS threshold $(\eta)$, (4.) Calculate precision, recall, and $\text{F}_1$ scores, (5.) Modify values of $(\gamma_1)$ and $(\gamma_2)$ and repeat steps (2-5), (6.) Choose the best values of $(\gamma_1)$ and $(\gamma_2)$ based on preference for precision, recall, and/or $\text{F}_1$ scores. It should also be noted that since there is no formal specification of the NMS threshold $(\eta)$, we default it to 0.5 for all experiments unless specified otherwise.

\section{Case Study: YOLOv2 using Darknet-19}\label{section:4}

While RMOPP is a suitable add-on for most of the available object detectors, we consider applying it to improve YOLOv2 in this case-study. YOLOv2 is one of the fastest existing object detectors; however, it detects many erroneous boxes and that makes it an excellent case-study for evaluating the efficacy of post-processing algorithms.

YOLOv2’s backbone is called as ‘Darknet-19’ that consists of 19 hidden convolutional layers, and it was first introduced in 2017 \cite{Redmon2017}. It is one of the fastest architecture around and performs about 5.58 billion operations per image \cite{Redmon2017}. The YOLOv2 loss function is a multi-part loss function which is a weighted sum of classification, localization, and confidence losses \cite{Redmon2016}. The ‘classification loss’ represents the loss based on the squared error of conditional class probabilities of the detected object given by the softmax layer of the detector. The ‘localization loss’ represents the loss based on the difference between coordinates of the bounding boxes for detected objects and the coordinates of bounding boxes in annotations. Note that a weighting parameter $({{\lambda }_{coord}})$ is used to increase the contribution of localization loss in the overall loss function. ‘Confidence loss’ represents the loss concerning the objectness score for each bounding box. Since we expect more boxes for background class, to eliminate the class imbalance, another weighting parameter $({{\lambda }_{noobj}})$ is used to down weight the contribution of confidence loss in the overall loss function when a bounding box is identified as a background object. $({{\lambda }_{coord}})$ and $({{\lambda }_{noobj}})$ are set to 5 and 0.5 by the Redmon \etal in \cite{Redmon2017}.

The MS-COCO dataset used for training YOLOv2 consists of around 118,000 images with an average of 7 boxes per image and a total of 80 unique class labels \cite{Lin2014}. For performance evaluation, we use the COCO \textit{minival} dataset which has 5000 images that were not used in training the YOLOv2, and set the input image resolution to 544x544 for all our experiments.

\subsection{Results: Precision, Recall, and $\bf \text{F}_1$ score}

In this case study, we use the definitions of precision, recall, and  $\text{F}_1$ score as given in Equations \ref{eq:8},\ref{eq:9} and \ref{eq:10}. A detection is considered as a true positive if the $(\text{IoU})$ is greater than 0.5 with at least one ground truth annotation of the same class label. Note that if multiple objects have $(\text{IoU})$ more than 0.5, then only the one with the highest $(\text{IoU})$ is selected as a true positive while others are considered to be false positives. All unmatched objects in the list of ground truths are considered to be false negatives. For this case-study $(\gamma_1)$ is varied between 1 and 10 with increments of 0.5 and $(\gamma_2)$ is varied between 0.1 and 1 with increments of 0.05. Beyond those limits, YOLOv2 showed extremely poor precisions or recalls.

\begin{equation} \label{eq:8}
\text{Precision = }\frac{\text{True Positives}}{\text{True Positives + False Positives}} 
\end{equation}

\begin{equation} \label{eq:9}
\text{Recall = }\frac{\text{True Positives}}{\text{True Positives + False Negatives}}
\end{equation}

\begin{equation} \label{eq:10}
{{\text{F}}_{1}}\text{ score = }\frac{\text{2 x Precision x Recall}}{\text{Precision + Recall}}
\end{equation}

\begin{figure}%
\centering
\subfloat[\label{fig:fig4a} The labels inside the figure denote recall while the color represents precision. The highlighted region encloses combinations of $(\gamma_1)$ and $(\gamma_2)$ where recall=0.5.]{{\includegraphics[width=0.4\textwidth]{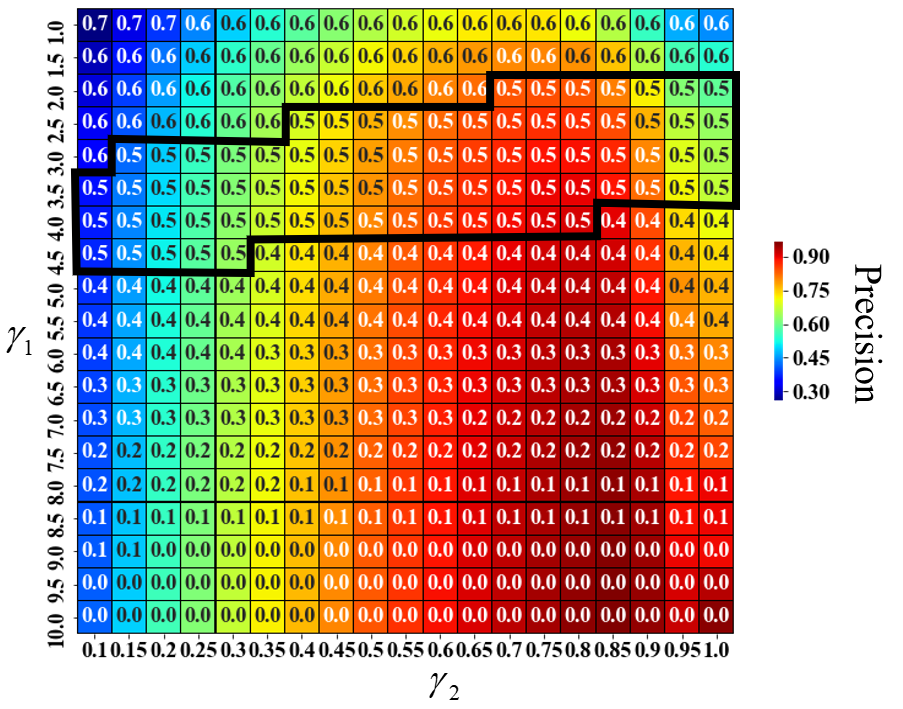} }}%

%\hspace{0.75em}%
\subfloat[\label{fig:fig4b} The labels inside the figure denote precision while the color represents recall. The highlighted region encloses combinations of $(\gamma_1)$ and $(\gamma_2)$ where precision=0.5.]{{\includegraphics[width=0.4\textwidth]{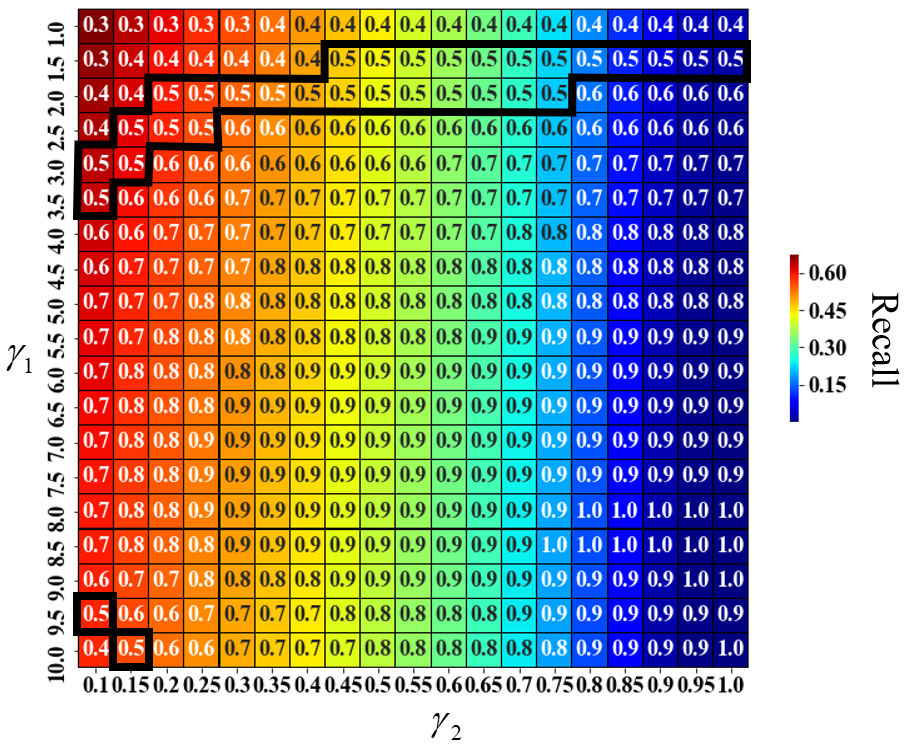} }}%

%\hspace{0.75em}%
\subfloat[\label{fig:fig4c} The highlighted region encloses combinations of $(\gamma_1)$ and $(\gamma_2)$ where $\text{F}_1$ score$\ge$0.5.]{{\includegraphics[width=0.4\textwidth]{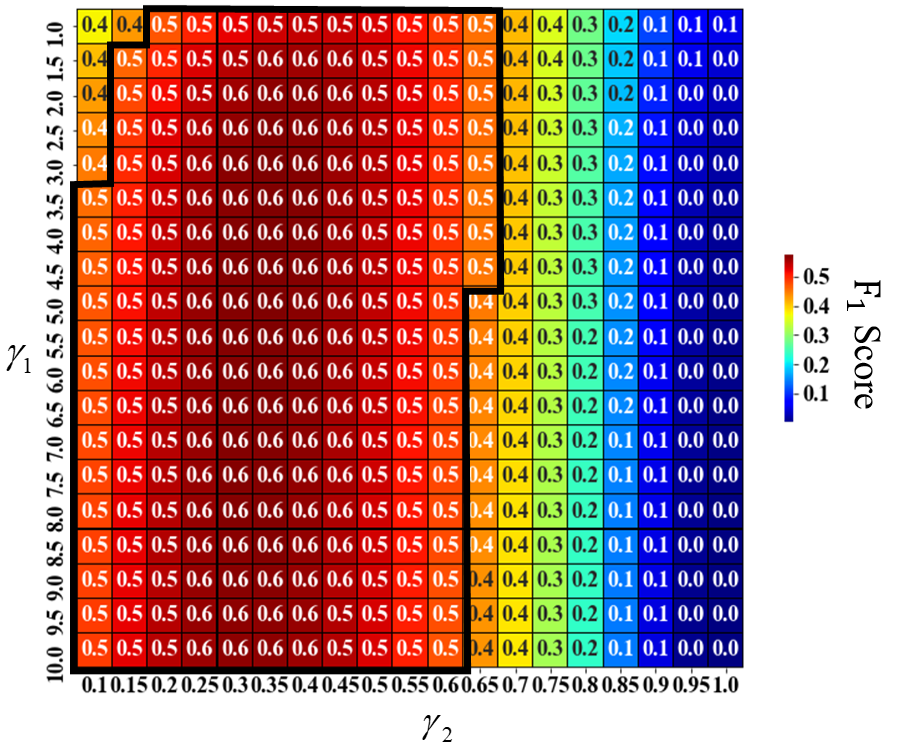} }}%
\caption{\label{fig:fig4}Effects of $(\gamma_1 )$ and $(\gamma_2 )$ on Precision, Recall and $\text{F}_1$ score.}%
\end{figure}

Changes in $(\gamma_1)$ and $(\gamma_2)$ have a significant effect on precision, recall, and $\text{F}_1$ scores. Essentially, $(\gamma_1)$ and $(\gamma_2)$ can be perceived as tuning knobs that can be used to tune the object detector for desired precision, recall, or $\text{F}_1$ score. In Figure \ref{fig:fig4}\subref{fig:fig4a} it is seen that as $(\gamma_1)$ is increased, a smooth decrease in recall is observed with recall almost dropping to 0 for the highest values of $(\gamma_2)$. Similarly, in Figure \ref{fig:fig4}\subref{fig:fig4b} it is observed that as $(\gamma_2)$ is increased, precision increases almost linearly to 0.9 or more when $(\gamma_2)$ is augmented from 0.1 to 1 for all values of $(\gamma_1)$ above 5.5. For values of $(\gamma_1)$ lower than 5.5 this increase is not so significant due to the correlation between $(\gamma_1)$ and $(\gamma_2)$.

When $(\gamma_1)$ and $(\gamma_2)$ are set to highest values, total detections drop, and a maximum precision condition (Recall$\approx$0) is realized. Similarly, when $(\gamma_1)$ and $(\gamma_2)$ are set to lowest values, total detections rise and a maximum recall condition (Precision$\approx$0) is realized. This helps build a sense of confidence which suggests that by employing finer adjustments in $(\gamma_1)$ and $(\gamma_2)$, we can explore the entire possible space of precision-recall offline. In this case study, multiple post-processed detectors with different $(\gamma_1)$ and $(\gamma_2)$ resulted in the same precision but different recall (and vice versa). The region highlighted in Figure \ref{fig:fig4}\subref{fig:fig4a} shows possible combinations of $(\gamma_1)$ and $(\gamma_2)$ where recall is fixed to 0.5. The regions highlighted in Figure \ref{fig:fig4}\subref{fig:fig4b} show possible combinations of $(\gamma_1)$ and $(\gamma_2)$ where precision is fixed to 0.5.

\begin{figure}
\begin{center}
\includegraphics[width=0.9\linewidth]{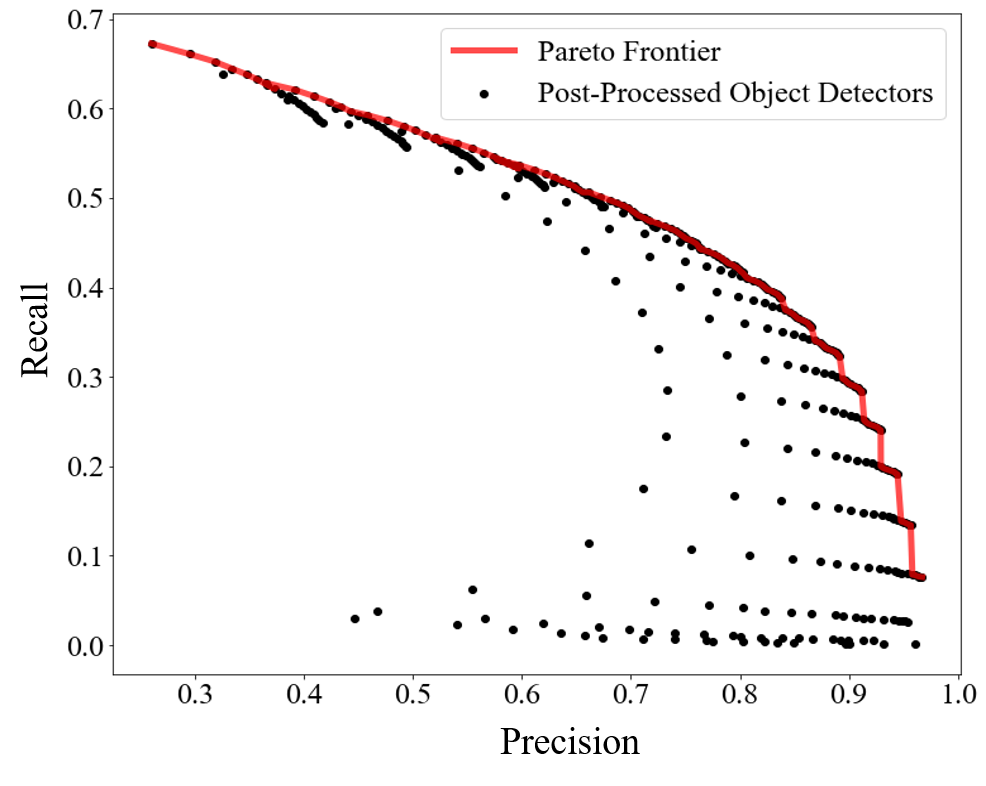}
\caption{Pareto Frontier based on precision and recall. Here, each combination of $(\gamma_1)$ and $(\gamma_2)$ is a uniquely post-processed object detector and is represented by a ‘black’ dot.}
\label{fig:fig5}
\end{center}
\end{figure}

Designing a bi-objective function to simultaneously optimize precision and recall is challenging, thus, researchers have developed metrics like $\text{F}_1$ score to address this challenge. In Figure \ref{fig:fig4}\subref{fig:fig4c} we plot a heat map of $\text{F}_1$ scores in Figure \ref{fig:fig4}\subref{fig:fig4c} by varying $(\gamma_1)$ and $(\gamma_2)$. Although Figure \ref{fig:fig4}\subref{fig:fig4c} helps us understand the complex correlations between $(\gamma_1)$, $(\gamma_2)$, precision and recall, it offers limited help in selecting values of $(\gamma_1)$ and $(\gamma_2)$ that maximize precision at given recall and vice-versa. Thus, we plot a Pareto frontier using results obtained by varying $(\gamma_1)$ and $(\gamma_2)$. The Pareto frontier is plotted as a function of precision and recall in Figure \ref{fig:fig5}. Here, we use the concepts of ‘Pareto Optimality’ \cite{Karuppusami2006} and extend them to the field of object detection for the first time, to the best of our knowledge. In object detection, ‘Pareto Optimality’ is defined as a condition where no improvements can be made to either precision or recall without some sacrifice in another. Such combinations of $(\gamma_1)$ and $(\gamma_2)$ which achieve an optimal trade-off between precision and recall are called ‘Pareto Optimal’ points. A set of such Pareto optimal points is called as ‘Pareto Frontier’. Pareto optimal points are said to dominate non-optimal points because there exists no other combination of $(\gamma_1)$ and $(\gamma_2)$ which achieves a better precision (or recall) at a higher recall (or precision). Figure \ref{fig:fig5} shows a plot of each unique combination of $(\gamma_1)$ and $(\gamma_2)$ and their associated precision and recall as a solid ‘black’ dot. A Pareto frontier is highlighted in ‘red’ in this figure which helps choose the best possible combinations of $(\gamma_1)$ and $(\gamma_2)$ that maximize precision at given recall or vice-versa. Figure \ref{fig:fig5}, thus, can be used to thoroughly explore and select a robust choice of hyper-parameters $(\gamma_1)$ and $(\gamma_2)$. 

Although Pareto Frontier helps in choosing the best combinations of $(\gamma_1)$ and $(\gamma_2)$, setting values for $(\gamma_1)$ and $(\gamma_2)$ that maximize precision (or recall) without ensuring some minimum recall (or precision) is not recommended as it will result in too few or too many detections, which is not practical. Thus, to avoid this, we set a minimum threshold on the $\text{F}_1$ score to be 0.5 in this case study. The optimal hyper-parameters achieving maximum precision, recall, and $\text{F}_1$ score under this condition are thus pinpointed using dominance principles in the Pareto frontier using Figure \ref{fig:fig5}. Table \ref{table:table2} summarizes these values.

\begin{table}
\centering
\begin{tabular}{|l|c|c|c|c|c|}
\hline
\begin{tabular}[c]{@{}c@{}}Maximizing\\Objective\end{tabular} &\begin{tabular}[c]{@{}c@{}}$\text{F}_1-$\\score\end{tabular}  & Recall & Precision & $\gamma_1$&$\gamma_2$\\
\hline\hline
Precision&0.50&0.35&0.88&10&0.55\\
Recall&0.50&0.60&0.43&3.5&0.15\\
$\text{F}_1$ score&0.57&0.49&0.68&4.5&0.35\\
\hline
\end{tabular}
\caption{\label{table:table2}Pareto Optimal $(\gamma_1)$ and $(\gamma_2)$ under different maximization objectives.}
\end{table}

\subsection{Comparison with Benchmark Methods}
\begin{table*}[!h]
\centering
\begin{tabular}{|l|c|c|c|c|c|c|c|c|}
\hline 
Method&Backbone&\begin{tabular}[c]{@{}c@{}}Post-Processing\\Method\end{tabular}&FPS&\begin{tabular}[c]{@{}c@{}}$\gamma_1$\end{tabular}&\begin{tabular}[c]{@{}c@{}}$\gamma_2$\end{tabular}&\begin{tabular}[c]{@{}c@{}}AP\\$\text[0.5:0.95]$\end{tabular}&\begin{tabular}[c]{@{}c@{}}AP\\$\text[0.5]$\end{tabular}&\begin{tabular}[c]{@{}c@{}}AP\\$\text[0.75]$\end{tabular}\\\hline\hline
YOLOv2 \cite{Redmon2017}&Darknet-19   &NMS(Baseline)  &\bf40  &-  &-  &0.216  &0.440  &0.192  \\
MetanAnchorGS \cite{Yang2018}&Darknet-19   &NMS  &- &- &- &0.212  &0.395&-  \\
Faster R-CNN \cite{Hu2018}&ResNet-50  &Soft-NMS  &9 &-  &-  &0.300  &0.523  &0.305  \\
Faster R-CNN \cite{Hu2018}&ResNet-50  &RelationNets  &17  & -&-  &0.303  &0.519  &0.315  \\
Faster R-CNN \cite{Tan2019}&ResNet-50-FPN  &Rank-NMS  &8  &-  &-  &\bf0.386  &\bf0.582  &\bf0.417  \\
YOLOv2&Darknet-19   &RMOPP-Best AP  &\bf40  &1  &0.1  &0.278  &\bf0.524  &0.266  \\
YOLOv2&Darknet-19   &RMOPP-Best Precision &\bf40  &10  &0.55  &0.216  &0.366  &0.229  \\
YOLOv2&Darknet-19   &RMOPP-Best Recall  &\bf40  &3.5  &0.15  &0.268  &0.502  &0.258  \\
YOLOv2&Darknet-19  &RMOPP-Best $\text{F}_1$ score  &\bf40  &4.5  &0.35  &0.253  &0.459  &0.251\\
\hline
\end{tabular}
\caption{\label{table:table3}Comparison of RMOPP across benchmarks using the COCO metrics.}
\end{table*}

For benchmark comparisons, we utilize the COCO AP metric that has been extensively used to evaluate various state-of-art detectors on the COCO challenge \cite{Lin2014}. This metric is based on the PASCAL VOC 2012 metric which used a similar strategy to evaluate object detectors \cite{M.2010}. Both VOC 2012 and COCO metrics calculate the per class Average Precision or (AP) by calculating the area under the Precision-Recall curve. This procedure is repeated for different $(\text{IoU})$ between 0.5 and 0.95. The individual AP for every $(\text{IoU})$ is reported as AP [IoU] and the mean AP is reported as AP [0.5:0.95].

\begin{figure}
\centering
\begin{subfigure}[t]{0.5\textwidth}
\centering
{\includegraphics[height=2.2in,width=0.95\linewidth]{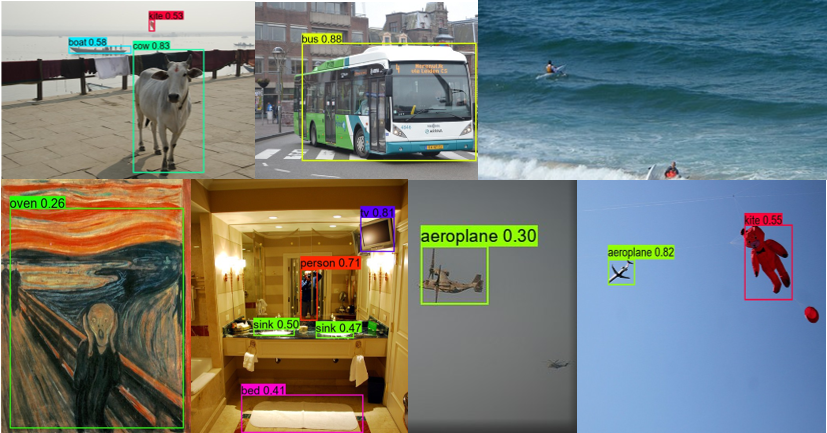} }
\caption{\centering YOLOv2 + Baseline}
\end{subfigure}
\hfill
\begin{subfigure}[t]{0.5\textwidth}
\centering
{\includegraphics[height=2.2in,width=0.95\linewidth]{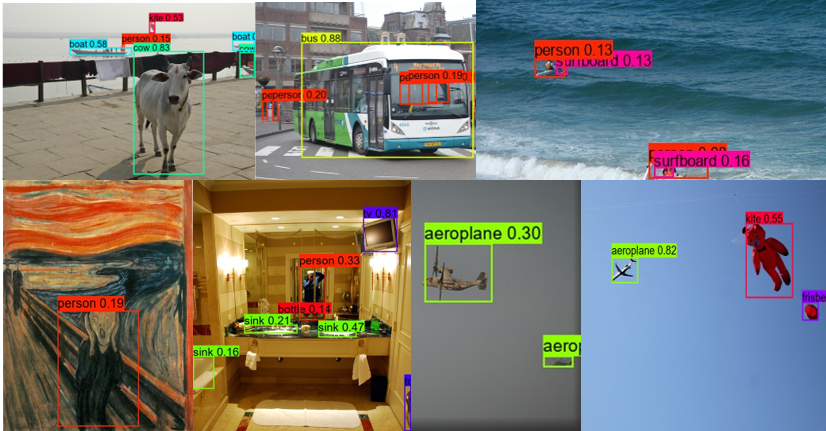} }
\caption{\centering YOLOv2 + RMOPP}
\end{subfigure}
\caption{\label{fig:fig6}YOLOv2+RMOPP shows significant improvement in results.}
\end{figure}

\begin{figure}[!h]
\centering
\begin{subfigure}[t]{0.5\textwidth}
\centering
{\includegraphics[height=2in,width=0.95\linewidth]{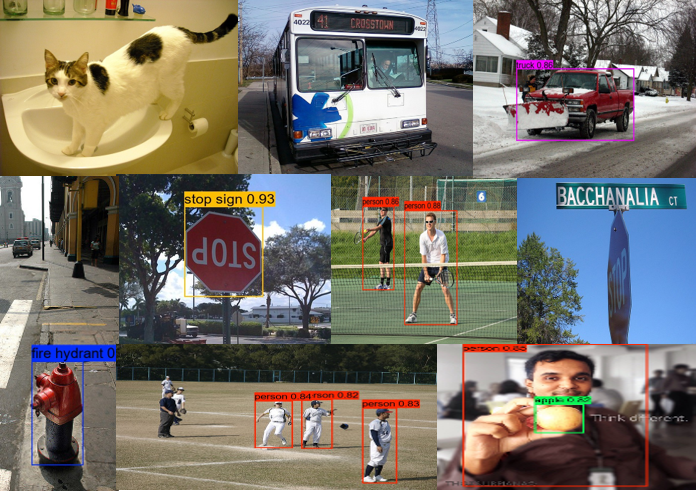} }
\caption{\centering YOLOv2 + Baseline}
\end{subfigure}
\hfill
\begin{subfigure}[t]{0.5\textwidth}
\centering
{\includegraphics[height=2in,width=0.95\linewidth]{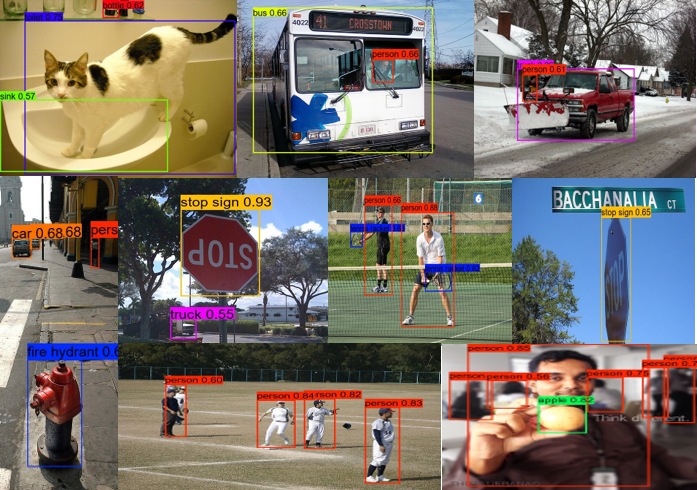} }
\caption{\centering YOLOv2 + RMOPP}
\end{subfigure}
\caption{\label{fig:fig7}YOLOv2+RMOPP shows significant improvement in results.}
\end{figure}

Table \ref{table:table3} summarizes the comparisons between the results of RMOPP and other benchmarked post-processing methods. Using Table \ref{table:table3}, it can be observed that RMOPP under different optimal settings significantly improves the COCO AP metrics for YOLOv2 without any loss in inference times (FPS). Specifically, for AP [0.5:0.95], RMOPP shows a $29\%$ increase compared to the original benchmark of YOLOv2 \cite{Redmon2017}. Henceforth, we mainly focus on AP [0.5] because for practical applications the detected boxes must have at least a $50\%$ intersection-over-union with the ground truth or it will be considered a bad detection. Using Table \ref{table:table3}, it can be observed that, the proposed method easily surpasses the performance by other benchmark post-processing methods such as Soft-NMS \cite{Bodla2017} and RelationNets \cite{Hu2018} while being significantly faster on an NVIDIA Titan X GPU. Here it should be noted that although Rank-NMS \cite{Tan2019} achieves better performance, it is significantly slower than the proposed method.  Figures \ref{fig:fig6} and \ref{fig:fig7} compares performance of YOLOv2 with traditional filtering and RMOPP. The compelling evidence in these figures helps prove the effectiveness of RMOPP over traditional algorithms.
\subsection{Comparison to State-of-Art Object Detectors}

Table \ref{table:table4} aims to compare the performance of the YOLOv2 with RMOPP to other state-of-art object detectors \cite{He2016,Lin2020,Sun2020}. The AP [0.5] results in Table \ref{table:table4} show that RMOPP improves the YOLOv2 performance to become comparable to that of its slower but more complex successor- YOLOv3. It also shows that better post-processing can significantly improve the performance of single-stage detectors like YOLOv2 to match that of a two-stage detector like Faster R-CNN.

\begin{table}
\centering
\begin{tabular}{|l|c|c|c|}
\hline
Method&$\gamma_1$ & $\gamma_2$   & \begin{tabular}[c]{@{}c@{}}AP\\ {[}0.5{]}\end{tabular} \\\hline\hline
\multicolumn{4}{|c|}{\textbf{Two-Stage Detectors}}\\
Faster R-CNN+++ \cite{He2016}& -   & -    & 0.557 \\
Faster R-CNN w FPN \cite{Lin2017}& -   & -    & 0.591 \\
Faster R-CNN w D-RMI \cite{Huang2017a}& -   & -    & 0.555 \\ 
Faster R-CNN w TDM \cite{Shrivastava2016}& -   & -    & 0.577\\\hline\hline
\multicolumn{4}{|c|}{\textbf{Single-Stage Detectors}}\\
YOLOv3 608x608 \cite{Farhadi2018}& -   & -    & 0.579\\
SSD513 \cite{Fu2017}& -   & -    & 0.504 \\ 
DSSD513 \cite{Fu2017}& -   & -    & 0.533 \\ 
RetinaNet+ResNet-101-FPN \cite{Lin2020}& -   & -    & 0.591\\ 
RetinaNet+ResNeXt-101-FPN \cite{Lin2020}& -   & -    & \bf 0.611\\ 
CenterNet+ResNet-18 \cite{Sun2020}& -   & -    & 0.466\\ 
OneNet+ResNet-18 \cite{Sun2020}& -   & -    & 0.457 \\
YOLOv2+RMOPP-Best AP & 1   & 0.1  & \bf 0.524\\ 
YOLOv2+RMOPP-Best Recall& 3.5 & 0.15 &\bf 0.502 \\ 
YOLOv2+RMOPP-Best $\text{F}_1$ score& 4.5 & 0.35 & \bf 0.459 \\ 
YOLOv2+RMOPP-Best Precision& 10  & 0.55 & \bf 0.366\\\hline 
\end{tabular}
\caption{\label{table:table4}Comparison of RMOPP with state-of-art.}
\end{table}

\section{Conclusion and Futurework}\label{section:5}
This paper introduces RMOPP: A robust multi-objective post-processing algorithm that improves the performance of pre-trained object detectors with a negligible impact on speed. Unlike existing uni-objective post-processing methods, the proposed algorithm allows for simultaneous optimization of precision and recall in object detection pipelines. When applied to YOLOv2, RMOPP showed a $29\%$ improvement in average precision to match similar performances of slower but more complex object detectors like YOLOv3 and Faster-RCNN. This work presents the first known usage of Pareto frontiers for post-processing hyper-parameter optimization in the field of object detection, to the best of our knowledge. A Bayesian post-processing technique may be feasible in future work to update the post-processing hyper-parameters for every inputted image and reduce training bias.

%-------------------------------------------------------------------------
{\small
\bibliographystyle{ieee_fullname}
\bibliography{objectdetection}
}

\end{document}